%% file: acl.tex
\pdfoutput=1

\documentclass[11pt]{article}

\usepackage{acl}

\usepackage{times}

\usepackage[T1]{fontenc}

\usepackage[utf8]{inputenc}

\usepackage{microtype}

\usepackage{graphicx}
\usepackage{blindtext}
\usepackage{multirow}
\usepackage{multicol}
\usepackage{float}
\usepackage{textcomp}
\usepackage{xcolor}
\usepackage{colortbl}
\usepackage{latexsym}
\usepackage{enumitem}
\usepackage{subfig}
\usepackage{wrapfig}
\usepackage{comment}
\usepackage{booktabs}
\usepackage{makecell}
\usepackage{tablefootnote}
\usepackage{multirow}
\usepackage{kantlipsum}
\usepackage{soul}

\usepackage[space]{grffile}

\definecolor{maroon}{RGB}{204,0,0}
\definecolor{green1}{RGB}{56,118,29}

\title{Improving Compositional Generalization with Self-Training for Data-to-Text Generation}

\author{Sanket Vaibhav Mehta$^{1}$\thanks{\hspace{1mm} Work performed during an internship at Google.} \quad Jinfeng Rao$^{2}$ \quad Yi Tay$^{3}$ \\ 
\textbf{Mihir Kale}$^{2}$ \quad \textbf{Ankur P. Parikh}$^{3}$ \quad \textbf{Emma Strubell}$^{1,3}$ \\\\
    $^{1}$Carnegie Mellon University, $^{2}$Google, $^{3}$Google Research \\
    \texttt{\{svmehta, estrubel\}@cs.cmu.edu} \\ \texttt{\{jinfeng, yitay, mihirkale, aparikh\}@google.com}
    }

\begin{document}
\maketitle
\begin{abstract}

Data-to-text generation focuses on generating fluent natural language responses from structured meaning representations (MRs). Such representations are compositional and it is costly to collect responses for all possible combinations of atomic meaning schemata, thereby necessitating few-shot generalization to novel MRs.
In this work, we systematically study the compositional generalization of the state-of-the-art T5 models in few-shot data-to-text tasks. We show that T5 models fail to generalize to unseen MRs, and we propose a template-based input representation that considerably improves the model's generalization capability.
To further improve the model's performance, we propose an approach based on self-training using fine-tuned BLEURT for pseudo-response selection.
On the commonly-used SGD and Weather benchmarks, the proposed self-training approach improves tree accuracy by $46\%+$ and reduces the slot error rates by $73\%+$ over the strong T5 baselines in few-shot settings. \footnote{Our code and data is available at \href{https://github.com/google-research/google-research/tree/master/compgen_d2t}{github.com/google-research/google-research/tree/master/compgen\_d2t}}

\end{abstract}

\input{sections/01_introduction}

\input{sections/02_experimentdesign}

\input{sections/03_selftraining}

\input{sections/04_results}

\input{sections/05_relatedwork}

\input{sections/06_conclusion}

\input{sections/07_ethics_statement}

\bibliography{custom}
\bibliographystyle{acl_natbib}

\newpage
\appendix
\input{sections/08_appendix}

\end{document}

%% file: sections/01_introduction.tex
\section{Introduction}

Data-to-text generation~\cite{duvsek2020evaluating,shen2020neural} is a critical component in today's task-oriented dialog systems for producing fluent natural language responses to users' requests. The task takes structured meaning representations (MRs) as input for natural language text response generation. Such representations are compositional, which allows for the combination of atomic meaning units in various ways to express the rich semantics encoded in languages. Recently, large pre-trained language models (LMs) have shown impressive results on many language understanding and generation tasks ~\citep{howard2018universal, peters2018deep, devlin2019bert, raffel2020exploring}, however it remains unclear how well these LMs generalize compositionally to novel semantic representations.

\begin{figure}
\includegraphics[trim=40 0 10 10,clip,width=0.53\textwidth]{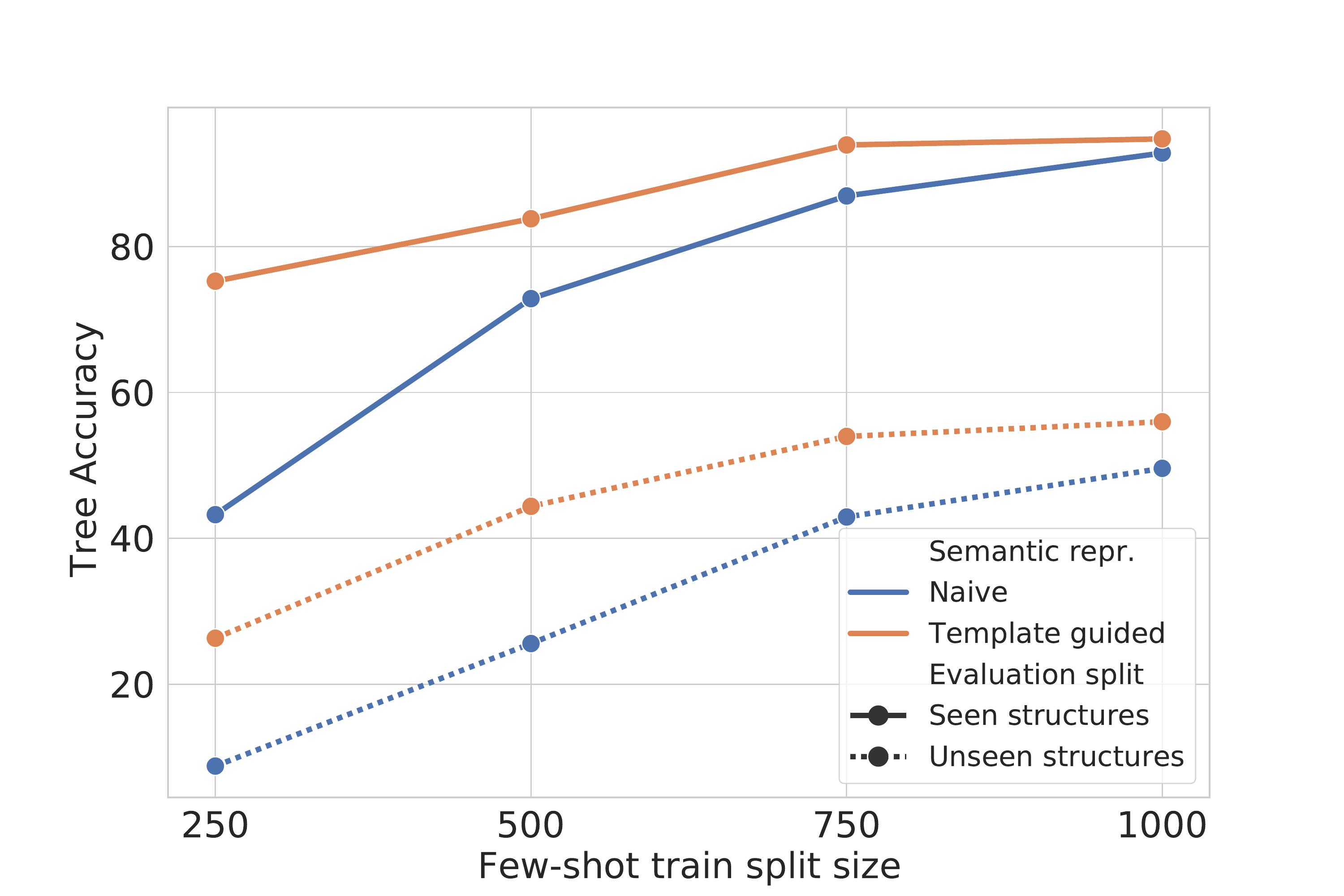}
\caption{Performance comparison (tree accuracy) between different few-shot splits and semantic representations. T5-small undergoes a significant drop in performance on the unseen split and our template-guided representation improves generalization, reducing the gap.}
\label{fig:fewshotcompgen}
\end{figure}

\begin{figure*}
\centering
\subfloat[Naive Structured Input]{
  \includegraphics[trim=220 120 250 120,clip,width=0.32\textwidth]{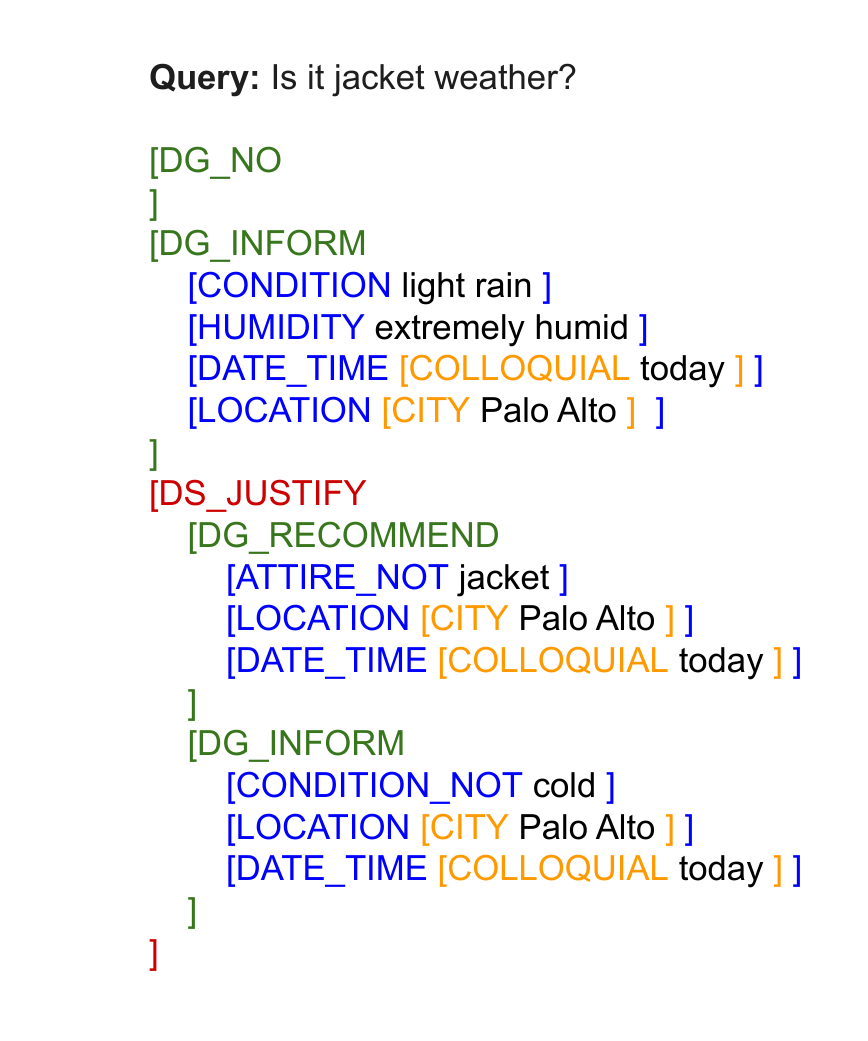}
}
\subfloat[Template Guided Structured Input]{
  \includegraphics[trim=220 120 250 120,clip,width=0.32\textwidth]{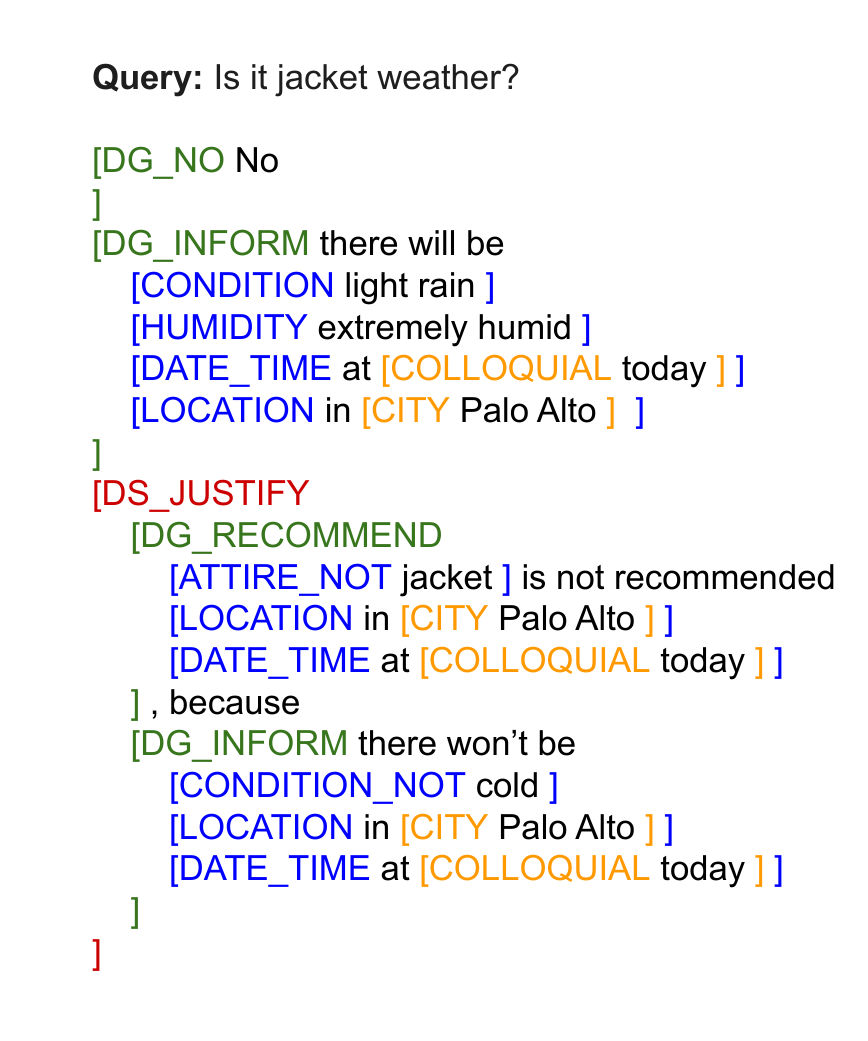}
}
\subfloat[Structured Target Response]{
  \includegraphics[trim=220 120 250 120,clip,width=0.32\textwidth]{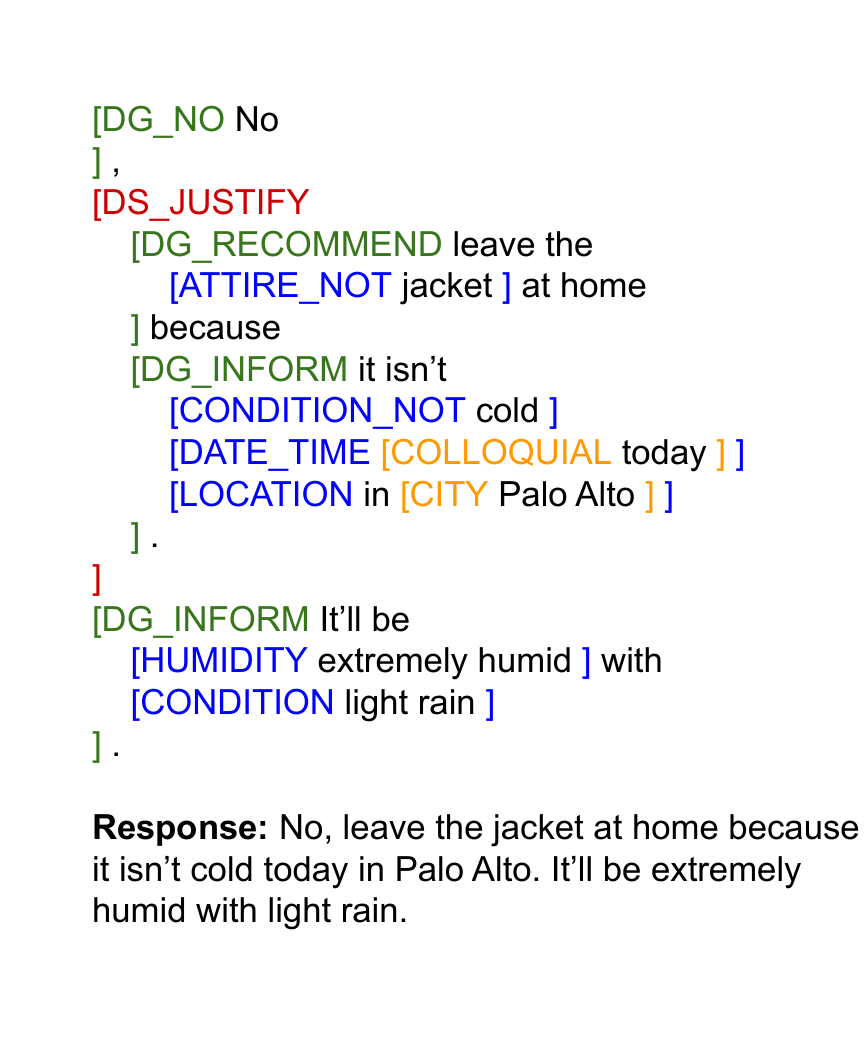}
}
\caption{Example compositional meaning representations ({\color{maroon} discourse relations}, {\color{green1} dialog acts}, {\color{blue} arguments}) \citep{balakrishnan2019constrained} - (a) naive input, (b) template guided input, and (c) structurally annotated target response.}
\label{fig:compmr}
\end{figure*}

There have been many studies revealing that large LMs often memorize the patterns from training data, while generalizing poorly to novel patterns. Compositionality in languages~\cite{banarescu2013abstract,konstas2017neural} further aggravates such issues as the number of novel structural combinations exponentially increases with the number of atomic semantic units. In recent years, we have seen progress on benchmarking and measuring compositional generalization for languages~\cite{andreas2019measuring}, from perspectives including specialized architectures \citep{lake2019compositional,rao2019tree} and learning strategies \citep{andreas2020good, akyurek2020learning}. However, most of these works study the generalization for NLU tasks like question answering \citep{keysers2019measuring} and semantic parsing \citep{kim2020cogs}. 
To the best of our knowledge, compositional generalization for natural language generation is still an under-explored problem, which is the focus of this work.

To answer the question of whether pre-trained LMs still suffer from lack of compositional generalization, we start with an empirical evaluation of T5~\citep{raffel2020exploring}, the state-of-the-art model on data-to-text generation tasks \citep{kale2020text}. In our study, we use the Weather dataset~\citep{balakrishnan2019constrained} consisting of tree-structured compositional MRs along with tree-structured output responses (see Figure~\ref{fig:compmr} for (a) naive MR and (c) target response). For evaluation, we compute the tree accuracy \citep{balakrishnan2019constrained} which measures exact match between input and generated tree-structures. 
In this study we observe $47\%$-$80\%$ (across different few-shot train splits) drop in the tree accuracy when evaluated on validation splits containing unseen tree-structures in comparison to splits containing seen tree-structures (Figure~\ref{fig:fewshotcompgen}). Furthermore, simply increasing the model size from T5-\emph{small} to T5-\emph{large} does not close the generalization gap (Table~\ref{tab:modelscale}), affirming our hypothesis that even strong seq-to-seq LMs fail to generalize compositionally. 

Inspired by~\citet{kale2020template}, we examine whether template-guided MRs are effective over naive MRs for tackling compositional generalization in data-to-text tasks. We introduce a simple template engine that traverses the compositional MR in a top-down manner and converts it to a text representation (Figure~\ref{fig:compmr}(b)). We hypothesize that such a template-guided setup reduces the change in representation between LM pre-training and fine-tuning. With template-guided MRs, we report up to $2$x increase in the tree accuracy over naive MRs on the validation split with unseen structures, demonstrating improved model generalization. 

\begin{table*}[h!]
     \small
     \begin{tabular}{lll}
     \toprule
       ID & Template Name & Template Body  \\
     \midrule
       1 & DG\_NO & {\color{green1}[DG\_NO} No {\color{green1}]} \\
       2 & DS\_JUSTIFY & {\color{maroon}[DS\_JUSTIFY} DG\_RECOMMEND, because DG\_INFORM {\color{maroon}]} \\
       3 & DG\_INFORM & \emph{IsSet}(\$condition) ? DG\_INFORM\_CONDITION \\
       & & : DG\_INFORM\_CONDITION\_NOT \\
       4 & DG\_INFORM\_CONDITION & {\color{green1}[DG\_INFORM} there will be {\color{blue}[CONDITION} \$condition {\color{blue}]} \\ 
       & & \emph{Optional}({\color{blue}[HUMIDITY} \$humidity {\color{blue}]}) DATETIME\_AND\_LOCATION {\color{green1}]} \\
       5 & DG\_INFORM\_CONDITION\_NOT & {\color{green1}[DG\_INFORM} there won't be {\color{blue}[CONDITION} \$condition {\color{blue}]} \\
       & & DATETIME\_AND\_LOCATION {\color{green1}]} \\
       6 & DATETIME\_AND\_LOCATION & \emph{Optional}(at {\color{blue}[DATE\_TIME} \$date\_time {\color{blue}]}) \emph{Optional}({in \color{blue}[LOCATION} \$location {\color{blue}]})  \\
       7 & DG\_RECOMMEND & {\color{green1}[DG\_Recommend} {\color{blue}[ATTIRE\_NOT} \$attire {\color{blue}]} is not recommended \\
        & & DATETIME\_AND\_LOCATION {\color{green1}]} \\
       
     \bottomrule
     \end{tabular}
    \caption{Example templates to convert a naive MR, Figure \ref{fig:compmr}(a), to template guided text representation, Figure \ref{fig:compmr}(b). A template could invoke other templates or some utility functions. The utility function \emph{IsSet} denotes whether the argument is set, and function \emph{Optional} returns empty text if the argument is not set. } 
    \label{tab:templates}
\end{table*}

We also propose to self-train the generation model to further boost performance by mitigating data sparsity in the low-data regime without requiring additional manual annotation. Concretely, we augment the limited labeled MRs with unlabeled novel MRs to iteratively bootstrap the model. To filter out noisy pseudo responses during self-training, we repurpose BLEURT~\cite{sellam2020bleurt}, a learned metric, to be a quality estimator. We synthetically generate datasets for finetuning BLEURT with the goal of identifying hallucinations, missing slot-values, and ungrammatical responses. 
In sum, our overall approach improves the tree accuracy on unseen structures of the FewShotWeather dataset by $12.3\%$-$46.4\%$ over strong T5 baselines. On unseen schemata of the FewShotSGD dataset, we reduce the slot error rate by $54.4\%$-$73.0\%$.

%% file: sections/02_experimentdesign.tex
\section{Case Study: Compositional Generalization in Data-to-Text Tasks}
\label{sec:study}
In this section, we are interested in investigating the following with respect to data-to-text tasks:
\begin{enumerate}[leftmargin=1cm,itemsep=0ex]
    \item [(Q1)] Do current state-of-the-art generation models compositionally generalize?
    \item [(Q2)] What is an effective semantic representation for tackling compositional generalization? 
    \item [(Q3)] Does scaling model size (and training data) trivially solve compositional generalization?
\end{enumerate}

\paragraph{Problem Setup} Data-to-text generation is the task of generating natural language text $y$ from meaning representation (MR) $x$. In the context of task-oriented dialog systems, the choice of MR ranges from a flat list of slot-value pairs \citep{duvsek2018findings} to a more expressive tree structure. \citet{balakrishnan2019constrained} defines \textbf{tree-structured MRs} consisting of arguments, dialog acts, and discourse relations, which we use in this work. They report significant gains in the naturalness of the generated responses with tree-structured MRs on the Weather domain dataset. Figure~\ref{fig:compmr} (a) visualizes an instantiation of such a tree-structured MR where the argument \textsc{location} is made up of a sub-argument (\textsc{city}), the dialog act \textsc{recommend} consists of three arguments (\textsc{attire\_not}, \textsc{location}, \textsc{date\_time}), and the discourse relation \textsc{justify} captures the relationship between two dialog acts (\textsc{recommend}, \textsc{inform}). 

We consider linearized versions of tree-structured MR $x$ and output response $y$. Generating the tree structure in the output enables us to compute the tree accuracy which helps to assess the structural correctness of the predicted response. 

\paragraph{FewShotWeather Dataset}
Due to the compositional nature of MRs, it is costly to collect responses for all combinations of discourse relations, dialog acts and arguments. In order to keep data labeling costs under control, we simulate a more realistic few-shot (or limited labeled data) setup. In the original Weather dataset, we have $25,390$ training examples spanning $4,690$ unique tree-structured MRs. An unique tree-structured MR is defined as a novel composition of discourse relations, dialog acts and argument names. Basically, they constitute non-terminals of a tree (Figure \ref{fig:compmr}(a) without terminals or argument values like extremely humid, light rain, today, Palo Alto, jacket, and cold). 

For the Weather dataset \citep{balakrishnan2019constrained}, we construct $4$ few-shot splits: 1shot-250, 1shot-500, 1shot-750, and 1shot-1000, where 1shot-$X$ denotes training split to include one example per unique tree-structured MR and in total $X$ unique tree-structured MRs.
Further, all $X$ examples in 1shot-$X$ are included while constructing  1shot-$Y$ splits, where $X < Y$.
We also make sure each discourse relation, dialog act and argument name is represented at least once in our few-shot splits. 
However, all combinations of these may not exist, thus allowing us to simulate structural shifts and evaluate compositional generalization. 
Based upon these splits, we construct two evaluation sets: \textit{seen} tree-structures (overlapping with tree-structured MRs from 1shot-250) and \textit{unseen} tree-structures (disjoint with tree-structured MRs from 1shot-1000) (see Section \ref{sec:datasetstats} for more details). Henceforth, all of the above splits constitute the \textit{FewShotWeather} dataset. We release these splits for future studies.

\subsection{Semantic Representation}

To answer (Q2), we use linearized tree structures as input to the T5 model ({\bf naive representation}). However, T5 based models are pre-trained on normal text as input, thereby creating a representation discrepancy between pre-training and fine-tuning. To alleviate this discrepancy, we introduce a simple template engine that recursively traverses the compositional MR in a top-down manner to generate a structure-aware text representation ({\bf template guided representation}). Some example templates to convert naive representation (Figure \ref{fig:compmr}(a)) to template guided representation (Figure \ref{fig:compmr}(b)) are listed in Table~\ref{tab:templates}. Each template, consisting of a name and a body, is invoked if a node in the MR (e.g., DG\_INFORM) matches its name. A template can also invoke other templates or some utility functions. For example, template 3 could invoke templates 4 or 5 based on the returned value of the utility function \emph{IsSet(\$condition)} (namely, whether the argument \$condition is set or not). 
Such a template engine requires developing only a linear number of templates with respect to the number of meaning units to convert a compositional MR to a text representation, without writing a template for each unique MR (4,690 unique MRs in the dataset).

In our study, we fine-tune the T5-small model using different few-shot train splits and report tree accuracy on validation splits. We observe that current state-of-the-art generation models undergo a significant drop in performance when evaluated on unseen tree structures. Specifically, with naive input representation, we observe $47\%$-$80\%$ (across different few-shot train splits) drop in tree accuracy, thus, providing evidence to answer (Q1) that the current model does not generalize to novel MRs.

On experimentation with template guided MRs and 1shot-250 train split, the tree accuracy on validation unseen split increases from 8.77 to 26.3 ($2$x increase over naive MRs), thus, answering (Q2) favorably (Figure~\ref{fig:fewshotcompgen}). However, across different few-shot train splits, template-guided MRs still undergo a significant $41\%$-$65\%$ drop in tree accuracy on the unseen split compared to the seen split. 

\subsection{Model scale}
Recent studies \citep{kaplan2020scaling, tay2021scale} show that model scale can affect the performance on several pre-training and downstream tasks. To understand how model scale affects the generalization to unseen structures, we consider three T5 variants: T5-small (77M), T5-base (120M),
and T5-large (800M). 
We fine-tune each of these models on the full training data (16,816 examples corresponding to 1000 unique tree-structured MRs from 1shot-1000 split) and convincingly answer (Q3): Increasing the model (and dataset) size does not close the performance gap between seen and unseen splits (Table~\ref{tab:modelscale}). Surprisingly, we observe that the T5-small model performs similarly or better than its larger counterparts. We use T5-small for the remaining experiments.

\begin{table}[t]
    \centering
    \small
    \begin{tabular}{l|r|r}
    \toprule 
    Model Size & Val. Seen & Val. Unseen \\
    \midrule
       T5-small (77M) & $99.54$ & $64.02$ \\
       T5-base (120M) & $99.63$ & $55.80$ \\
       T5-large (800M) & $99.36$ & $58.45$ \\
    \bottomrule
    \end{tabular}
    \caption{Performance comparison (tree accuracy) between different T5 model variants. Each T5 model is fine-tuned on full Weather dataset (16,816 examples) and evaluated on validation seen and unseen splits. We observe that increasing the model size does not close the compositional generalization gap.}
    \label{tab:modelscale}
\end{table}

%% file: sections/03_selftraining.tex
\section{Self-training}
\label{sec:stwithbleurt}

\begin{figure*}[t]
\centering
  \includegraphics[trim=0 150 0 120,clip,width=0.8\textwidth]{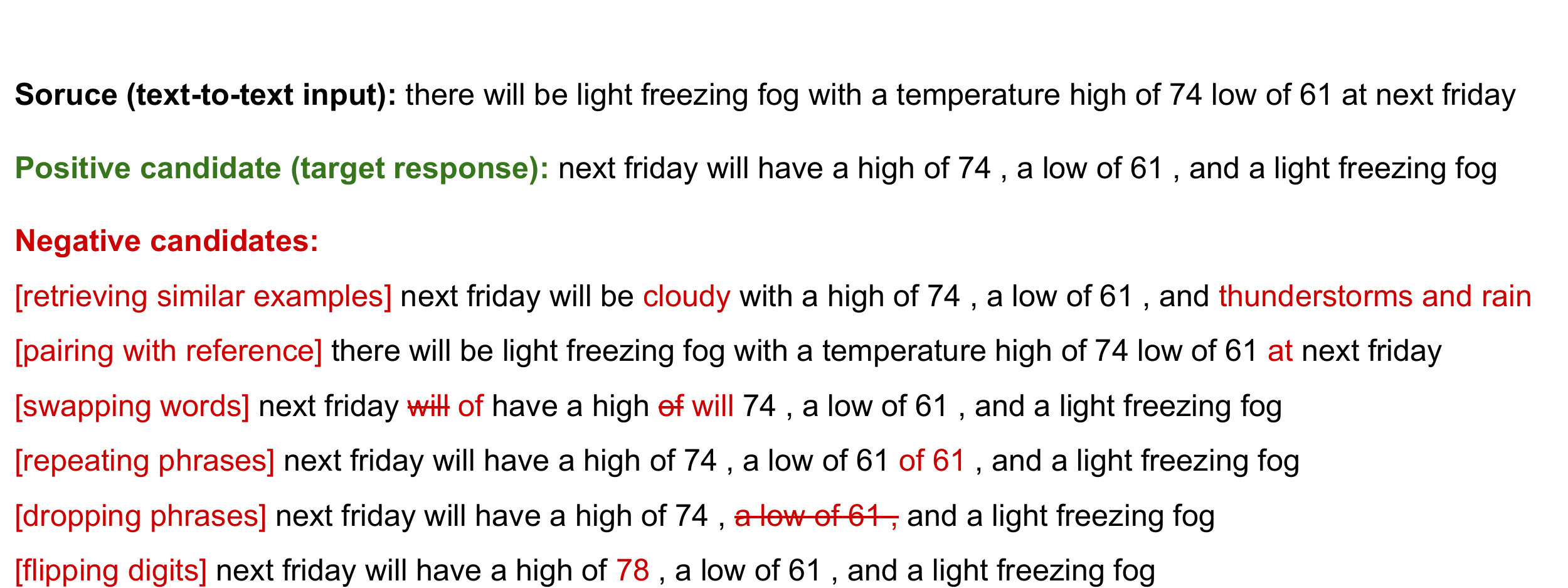}
\caption{Synthetically constructed positive and negative candidates for BLEURT fine-tuning.}
\label{fig:bleurt_ft_examples}
\end{figure*}

As discussed earlier, the compositional nature of MRs makes it difficult to collect responses for all combinations. However, with access to data simulators \citep{rastogi2020towards}, it is feasible to automatically generate large amounts of unlabeled MRs. Given limited labeled MRs, $S=\{x^i, y^i\}_{i=1}^n$, and assuming access to unlabeled MRs, $U=\{x^i\}_{i=1}^m$, we investigate self-training \citep{scudder1965probability}, a semi-supervised learning approach to effectively use $U$ to improve compositional generalization. 

Self-training starts from a model trained on labeled data $S$, iteratively applies the current model to generate pseudo-labels on unlabeled data $U$, and then re-trains the current model on the augmented version of $S$ and (subset of) $U$. For self-training to be effective, one needs to carefully select confident pseudo labels to alleviate the risk of reinforcing the model's mistakes \citep{he2019revisiting}. This issue gets further exacerbated in the context of generation tasks, where neural models are prone to hallucinate additional content not supported by the input \citep{maynez2020faithfulness}.

With recent developments in learned evaluation metrics that penalize the model for hallucination, fluency, etc., we pose the question: \textit{Can we repurpose those metrics to assess the quality of pseudo-responses during self-training?} Formally, given a pair of template guided MR (source) and model predicted response (candidate), we want a model that estimates the response quality by looking for hallucinations, fluency, coverage of argument value-pairs. Ideally, to learn such a model we require a large amount of positive and negative text pairs. To alleviate this requirement, we propose synthesizing the examples using the limited labeled task dataset. Furthermore, we initialize our quality estimation model using a pre-trained BLEURT \citep{sellam2020bleurt}, which is shown to be sample efficient and robust to data shifts as a learned evaluation metric.
Once we have a fine-tuned BLEURT model, we use it to select pseudo-responses using a selection threshold for self-training.

\subsection{Fine-tuning BLEURT}
We synthetically generate the dataset for fine-tuning BLEURT using the labeled dataset available for each of our experiments.
Template guided inputs and ground truth target responses are paired as positive examples (rating: $1.0$). 
We use the following transformations on the target responses to create negative examples (rating: $0.0$):

\noindent \textbf{Retrieving similar examples:} For every input $x$, we rank all other inputs from the dataset using the BLEU score and select top-k examples below a certain threshold ($90.0$). Target responses corresponding to these top-k examples are paired with $x$ to construct negative examples. Intuitively, these responses partially overlap with input $x$ in terms of the content and inform a fine-tuned model to handle hallucinations.

\noindent \textbf{Pairing with reference:} Template guided inputs need not be grammatically correct. Pairing the input $x$ with itself as a response provides grammatically incorrect negative examples.

\noindent \textbf{Swapping, repeating and dropping phrases, flipping digits:} Using these methods, we prepare a fine-tuned BLEURT for structurally inconsistent behaviors of the NLG system. 
Figure~\ref{fig:bleurt_ft_examples} visualizes an instantiation of different transformations to construct negative examples.

%% file: sections/04_results.tex
\begin{table*}[h!]
    \centering
    \small
    \begin{tabular}{l|c|rr|rr|p{1.0cm}|rr|rr}
    \toprule 
    Pseudo- & \multicolumn{5}{c|}{FewShotWeather} & \multicolumn{5}{c}{FewShotSGD} \\
    response & Train & \multicolumn{2}{c|}{Seen structures} & \multicolumn{2}{c|}{Unseen structures} & Train & \multicolumn{2}{c|}{Seen schemata} & \multicolumn{2}{c}{Unseen schemata} \\
    selection & split & BLEU $\uparrow$ & Tree & BLEU $\uparrow$ & Tree & split & BLEU $\uparrow$ & SER $\downarrow$ & BLEU $\uparrow$ & SER $\downarrow$ \\
    strategy & & & Acc. $\uparrow$ & & Acc. $\uparrow$ \\
    \midrule
    None & \multirow{3}{*}{1shot-250} & $69.16$ & $73.68$ & $50.40$ & $29.83$ & \multirow{3}{*}{\parbox{1.0cm}{5-shot (558)}} & $20.66$ & $22.84$ & $20.52$ & $19.93$ \\
    Vanilla & & $69.25$ & $73.77$ & $51.87$ & $31.37$ & & $23.03$ & $15.15$ & $21.97$ & $15.96$ \\
    BLEURT & & \textbf{69.59} & \textbf{84.12} & \textbf{52.34} & \textbf{43.68} & & \textbf{25.22} & \textbf{4.78} & \textbf{24.13} & \textbf{5.39} \\
    \midrule
    None & \multirow{3}{*}{1shot-500} & $69.40$ & $83.59$ & $53.62$ & $46.58$ & \multirow{3}{*}{\parbox{1.0cm}{10-shot (1,075)}} & $21.45$ & $21.64$ & $22.79$ & $14.98$ \\
    Vanilla & & $68.75$ & $89.21$ & $54.27$ & $49.91$ & & $23.50$ & $17.90$ & $24.38$ & $7.67$ \\
    BLEURT & & $68.19$ & \textbf{93.40} & \textbf{56.12} & \textbf{55.30} & & \textbf{25.63} & \textbf{4.29} & \textbf{25.49} & \textbf{3.82} \\
    \midrule
    None & \multirow{3}{*}{1shot-750} & $69.81$ & $92.86$ & $54.49$ & $54.02$ & \multirow{3}{*}{\parbox{1.0cm}{20-shot (2,140)}} & $22.84$ & $16.74$ & $25.14$ & $11.51$ \\
    Vanilla & & $73.02$ & $96.61$ & $54.32$ & $54.19$ & & $23.19$ & $14.92$ & $25.47$ & $9.11$ \\
    BLEURT & & $72.00$ & \textbf{97.23} & \textbf{55.21} & \textbf{58.89} & & \textbf{26.63} & \textbf{3.33} & \textbf{27.38} & \textbf{3.77} \\
    \midrule
    None & \multirow{3}{*}{1shot-1000} & $72.89$ & $95.18$ & $53.97$ & $55.64$ & \multirow{3}{*}{\parbox{1.0cm}{40-shot (4,312)}} & $25.72$ & $7.60$ & $26.52$ & $5.97$ \\
    Vanilla & &  $73.38$ & $96.16$ & $55.04$ & $60.09$ & & $26.65$ & $5.00$ & $26.61$ & $4.20$ \\
    BLEURT & & \textbf{73.82} & \textbf{98.48} & \textbf{57.11} & \textbf{62.48} & & \textbf{27.48} & \textbf{2.37} & \textbf{27.53} & \textbf{2.72} \\
    \midrule
    Full & 16,816 & $74.43$ & $99.55$ & $62.44$ & $65.47$ & 164,978 & $29.28$ & $1.12$ & $28.76$ & $1.54$ \\
    \bottomrule
    \end{tabular}
    \caption{Comparing performance in terms of BLEU, tree accuracy (Tree Acc.), and slot error rate (SER) between vanilla and BLEURT based pseudo-response selection strategies on FewShotWeather and FewShotSGD test splits. All results are for the T5-small model with template guided input representation. Pseudo-response selection strategy \emph{None} denotes fine-tuned T5-small baseline without self-training. $\uparrow$ indicates higher is better, $\downarrow$ indicates lower is better. Overall, BLEURT based self-training improves the performance on (un)seen structures/ (un)seen schemata over vanilla self-training.
    }
    \label{tab:selftrain_results}
\end{table*}

\section{Experimentation}

\subsection{Datasets and Metrics}
\label{sec:datasetstats}
\paragraph{FewShotWeather}

The original Weather dataset \citep{balakrishnan2019constrained} has $25,390$ training examples. Each example consists of a user query, the tree-structured MR, the tree-structured annotated response and metadata. As discussed in Section~\ref{sec:study}, we create new canonical subsets for compositional generalization experiments, FewShotWeather with 1shot-250 (approx. $1\%$ of original training data), 1shot-500, 1shot-750, and 1shot-1000 splits. We repurpose all the remaining $24$k training examples as unlabeled examples for self-training. Our evaluation splits have $1,087/1,121$ (val/test) examples with seen tree-structures, and $1,095/1,170$ (val/test) examples with novel tree-structures. We report tree accuracy and BLEU-4 \citep{papineni2002bleu} for the FewShotWeather dataset. 

\paragraph{FewShotSGD}
The original multi-domain Schema Guided Dialogue (SGD) dataset \citep{rastogi2020towards} has $160$k examples spanning across $20$ domains (e.g., Banks, Travel, Weather, etc.). For each of these domains, there are different services with a total of $45$ different schemata. Schema here refers to the combination of intents and slots, which change with services and domains. Further, not all domains and services are observed during training. Therefore, we use this dataset to study generalization to unseen schemata. Specifically, we use the few-shot variant of the dataset, FewShotSGD, as introduced by \citet{kale2020template}. The FewShotSGD benchmark consists of $k$-shot splits (5/10/20/40), 
where $k$ denotes the number of dialogues selected per train domain. The few-shot train splits have 558/1,075/2,140/4,312 (5/10/20/40-shot) examples. Evaluation splits have 13,748/10,216 (val/test) examples with seen schema, and 10,386/26,568 (val/test) examples with novel schema. 
Following \citet{kale2020template}, we report BLEU-4 and slot error rate (SER) \citep{duvsek2019neural}. SER measures the fraction of examples where at least one slot was incorrectly copied from the input (lower SER is better). 

\subsection{Implementation}
For each of the experiments we fine-tune the off-the shelf T5.1.1.small checkpoint\footnote{\href{https://github.com/google-research/text-to-text-transfer-transformer/blob/main/released\_checkpoints.md}{github.com/google-research/text-to-text-transfer-transformer/blob/main/released\_checkpoints.md}}. It has 6 layers each in encoder and decoder with a total of $77$M parameters. We set the maximum sequence length to $512$, batch size to $16$ and a constant learning rate of $0.001$ for Adafactor optimizer \citep{shazeer2018adafactor}. All models are fine-tuned on a 4x4 TPU slice, each taking around 2-3 hours to finish $5000$ steps. We evaluate models after every $200$ steps and retain the checkpoint yielding best tree accuracy (for FewShotWeather) or BLEU (for FewShotSGD) on the held-out validation seen split. During inference, we set the beam size to $4$ and length penalty $\alpha = 0.6$. 

While constructing the fine-tuning dataset for BLEURT, we generate up to $4$ different negative candidates for each of the $6$ transformations. We upsample the positive examples to be half the total number of negative examples and retain random $10\%$ of total examples for validation set. For fine-tuning the BLEURT model, we start with publicly available BLEURT-20-D12 \citep{sellam2020bleurt}. We set the maximum sequence length to $512$, batch size to $32$, a learning rate 1e-6, and fine-tune for $100$k steps. We use the held-out validation set to select the best checkpoint for self-training.

\begin{table*}[h!]
    \centering
    \small
    \begin{tabular}{l|c|r|cc|cc}
    \toprule 
    Model & Self- & No. of & \multicolumn{4}{c}{FewShotWeather} \\
     & training & training & \multicolumn{2}{c}{Seen structures} & \multicolumn{2}{c}{Unseen structures} \\
     & iteration & examples & BLEU $\uparrow$ & Tree Acc. $\uparrow$ & BLEU $\uparrow$ & Tree Acc. $\uparrow$ \\
    \midrule
    Baseline & - & 250 & $69.16$ & $73.68$ & $50.40$ & $29.83$ \\
    \midrule
      \multirow{2}{*}{Vanilla} & 1 & + $14,742$ & $69.25$ & $73.77$ & $51.87$ & $31.37$ \\
       & 2 & + $4,170$ & $68.72$ & $73.06$ & $51.92$ & $31.11$ \\
      \midrule
      \multirow{2}{*}{BLEURT-250} & 1 & + $14,742$ & $69.64$ & $83.85$ & $52.10$ & $41.03$ \\
       & 2 & + $4,170$ & $69.59$ & $84.12$ & $52.34$ & $43.68$ \\
       \midrule
       \multirow{2}{*}{BLEURT-1000} & 1 & + $14,021$ & $70.95$ & $84.83$ & $52.13$ & $45.47$ \\
       & 2 & + $4,772$ & \textbf{70.47} & \textbf{85.64} & \textbf{53.08} & \textbf{47.44} \\
    \bottomrule
    \end{tabular}
    \caption{Model performance over multiple self-training iterations with FewShotWeather 1shot-250 train split. BLEURT-X denotes BLEURT model fine-tuned using 1shot-X train split. We observe that BLEURT model fine-tuned with larger datasets further enhances the self-training performance, especially on unseen structures.}
    \label{tab:multiselftrain}
\end{table*}

\subsection{Self-Training}

In this section, we compare the performance of BLEURT based pseudo-response selection strategy with that of \textit{vanilla} self-training. For each experiment, we \textit{randomly} sample an equal number of examples for vanilla self-training and the BLEURT model to explicitly control for the sample complexity. We run $3$ iterations of the self-training unless explicitly specified and set the BLEURT score selection threshold to $0.99$. We study the performance on a dataset (FewShotWeather) with tree-structured outputs as well as show the generality of our method on a dataset (FewShotSGD) without explicit tree-structured outputs. Note that naive T5 fine-tuning with template guided input representation constitutes a strong baseline for few-shot experiments as shown by \citet{kale2020template}. We include results from this baseline under \emph{None} pseudo-response selection strategy as it does not involve self-training.

\noindent \textbf{Unseen tree structures (FewShotWeather)}
Table~\ref{tab:selftrain_results} reports the performance of different methods as a function of the number of labeled examples.
We observe that the performance for all methods improves with more training data. Across all few-shot splits, we observe that BLEURT based self-training improves over vanilla self-training both in terms of tree accuracy and BLEU. Empirically, we see that relative gains in tree accuracy (over the T5-small baseline) from vanilla self-training are comparable on both unseen and seen splits (e.g., $7.15\%$ v.s. $6.72\%$, 1shot-500). On the other hand, BLEURT based self-training significantly improves the relative performance on the unseen split in comparison to seen splits (e.g.,  $18.72\%$ vs. $10.5\%$, 1shot-500), thus showcasing the effectiveness of selecting quality pseudo-responses for improving performance on unseen tree-structures.

\noindent \textbf{Unseen schema (FewShotSGD)}
Table~\ref{tab:selftrain_results} reports the performance on the FewShotSGD dataset. Similar to results on the FewShotWeather dataset, we observe that the performance improves with more training data. Further, the performance of the baseline T5-small model is comparable to seen and unseen schemata. These gains can be attributed to the benefits of using template guided MRs. In comparison to vanilla self-training, BLEURT based approach improves the overall performance across all few-shot splits on both seen and unseen schema. For example, with 5-shot experiments, BLEURT based selection strategy reduces the SER on unseen schema from 19.93 to 5.39 ($73\%$ improvement) in comparison to the baseline T5 model. On the other hand, vanilla self-training reduces the SER only by 3.97 ($20\%$), thus showcasing the effectiveness of the proposed approach in filtering pseudo-responses with missing slot-value pairs. These results confirm that BLEURT based self-training is a generic method and can be plugged in to existing methods to improve the few-shot generalization capabilities of existing SOTA generation models.

\noindent \textbf{Performance with respect to self-training iterations}
We iteratively self-train the model starting from a T5-small baseline and continue adding unlabeled examples up to $3$ iterations. From Table~\ref{tab:multiselftrain} and \ref{tab:multiselftrainsgd}, we see that model performance improves across the self-training iterations. However, the number of additional examples added decreases over iterations, thus suggesting that $2$-$3$ iterations might be enough to obtain benefits from self-training.

\noindent \textbf{Quality of fine-tuned BLEURT models}
For all our experiments, we use the few-shot labeled datasets for fine-tuning the BLEURT model. To investigate self-training performance with a BLEURT model fine-tuned on a large dataset, we set up an experiment on the FewShotWeather dataset, where we fine-tune the BLEURT model on a 1shot-1000 train split (BLEURT-1000) and use it for self-training with 1shot-250. From Table~\ref{tab:multiselftrain}, we see that self-training with BLEURT-1000 performs significantly better than BLEURT-250, especially on unseen structures, thereby confirming the intuition that self-training is sensitive to the quality of the BLEURT model.

\subsection{Human evaluation}
Aside from automatic metrics-based evaluation, we also perform a human evaluation study by asking annotators to assess the quality of the generated responses from different models. For each example, human annotators are shown user query, generated response and the ground truth response. 
They are asked to provide ratings on a scale of 1 (bad), 2 (slightly bad) to 3 (good) along two dimensions: \textit{grammaticality}, \textit{naturalness}, rating on a scale of 0 (less) to 1 (adequate) for \textit{informativeness}, and binary rating for \textit{accuracy}. 
Similar to \citep{balakrishnan2019constrained}, grammaticality evaluates the response for subject-verb agreement, repetitions, and grammatical completeness. Naturalness measures whether the response sounds coherent and natural by the response itself. Informativeness measures whether the response contains the right amount of relevant information to the user query and accuracy evaluates the response for hallucinations (incorrectly added slots), missing slots by comparing it against the reference. For each evaluation split (seen/unseen), we randomly select $200$ examples and collect ratings from $3$ different annotators. For the FewShotWeather/SGD datasets, we consider models trained with 1shot-250/5-shot splits and compare them with models fine-tuned on the full dataset. In total, we collect $7,200$ annotations, each with $3$ ratings. Table~\ref{tab:humaneval} reports results for human evaluation study.

\begin{table}[t]
    \centering
    \small
    \begin{tabular}{l|llll}
    \toprule 
    Model & Gram & Nat & Info & Acc\\
    \midrule
     \multicolumn{5}{l}{FewShotWeather (Seen split)} \\
      Baseline & 2.59 & 2.55 & \textbf{0.81} & 0.94\\
      BLEURT & \textbf{2.66$^{1}$} & \textbf{2.63$^{1}$} & 0.80 & 0.93\\
      Full & \textbf{2.66$^{1}$} & 2.61 & 0.80 & \textbf{0.95}\\
      \midrule
      \multicolumn{5}{l}{FewShotWeather (Unseen split)} \\
      Baseline & 2.43 & 2.41 & 0.75 & 0.79\\
      BLEURT & 2.50$^{1}$ & 2.46$^{1}$ & 0.76 & 0.80\\
      Full & \textbf{2.53$^{1}$} & \textbf{2.50$^{1}$} & \textbf{0.79$^{1}$} & \textbf{0.86$^{1,2}$} \\
      \midrule
      \multicolumn{5}{l}{FewShotSGD (Seen split)} \\
      Baseline & 2.72 & 2.66$^{2}$ & 0.79 & 0.76 \\
      BLEURT & 2.69 & 2.59 & \textbf{0.81} & 0.88$^{1}$ \\
      Full & \textbf{2.83$^{1,2}$} & \textbf{2.74$^{1,2}$} & \textbf{0.81} & \textbf{0.94$^{1,2}$} \\
      \midrule
      \multicolumn{5}{l}{FewShotSGD (Unseen split)} \\
      Baseline & 2.70 & 2.61 & 0.77 & 0.72 \\
      BLEURT & 2.67 & 2.60 & 0.79 & 0.86$^{1}$ \\
      Full & \textbf{2.83$^{1,2}$} & \textbf{2.73$^{1,2}$} & \textbf{0.82$^{1,2}$} & \textbf{0.94$^{1,2}$} \\
    \bottomrule
    \end{tabular}
    \caption{Human evaluation results comparing different models. Grammaticality (Gram), naturalness (Nat) are on the scale of $1$ to $3$, informativeness (Info) is on the scale of $0$ to $1$, and accuracy (Acc) is binary. The superscripts $1, 2, 3$ indicate that model is significantly better than baseline, BLEURT-based self-training, and model trained with full data, respectively, as determined by one-sided paired t-test with $p<0.05$.}
    \label{tab:humaneval}
\end{table}

\noindent \textbf{FewShotWeather} Similar to automatic metrics, we see a drop in human ratings on the unseen split (compared to seen split), confirming the model's lack of generalization to novel MRs. On both the evaluation splits, our approach outperforms the baseline model with significant results on grammaticality and naturalness ratings. Moreover, the responses from the self-trained model are comparable (in terms of the human ratings) with that of the model fine-tuned with the full dataset, demonstrating the effectiveness of our approach.

\noindent \textbf{FewShotSGD} Apart from generating natural responses, model responses must be factually grounded in the input data and address user queries. On FewShotSGD, we see that our approach significantly improves informativeness and accuracy rating over the baseline model. Surprisingly, we see a drop on naturalness when evaluated on seen schemata.

\subsection{Qualitative Analysis}
\label{sec:qualanalysis}
In Table~\ref{tab:qualitative} (and Tables~\ref{tab:qualitative_weather}, \ref{tab:qualitative_sgd} in Appendix~\ref{sec:appendix}) we visualize the sample responses generated using different models for unseen test splits. We consider three models: T5-small baseline, BLEURT based self-training, and model trained with full data. For the FewShotWeather/ FewShotSGD datasets, we consider models trained with 1shot-250/ 5-shot train splits. We see that the baseline model fails to generate responses that are coherent and factually grounded in the input. They hallucinate to generate novel concepts like \textit{cloudy hail}, drop relevant details like \textit{cafe located in Emeryville}, and are repetitive in nature. We also report the BLEURT score along with human ratings per sample and see that they are reflective of the response quality.
\begin{table*}[h!]
    \centering
    \small
    \begin{tabular}{p{1.4cm}rrrrrp{8.1cm}}
    \toprule 
    Fields & BLEURT & Gram & Nat & Info & Acc & Input or output response \\
    \midrule
    \textit{User query} & - & - & - & - & - & What will the temperature be tomorrow morning \\
    \textit{Template} & - & - & - & - & - & There will be temperatures between 76 and 80 tomorrow morning there will be partly cloudy tomorrow morning \\
    \textit{Reference} & - & - & - & - & - & The temperature for tomorrow morning will be between 76 and 80 fahrenheit along with partly cloudy skies \\
    \textbf{Predictions} \\
    \textit{Baseline} & -0.002 & 2.17 & 1.67 & 0.67 & 1.0 & Expect partly cloudy skies and tomorrow morning. {\color{red} Expect partly cloudy skies and tomorrow morning} \\
    \textit{Self-training} & 0.997 & 3.00 & 2.67 & 0.67 & 1.0 & Expect it to be between 76 and 80 and partly cloudy tomorrow morning \\
    \textit{Full} & 0.995 & 2.83 & 2.83 & 0.50 & 1.0 & The weather for tomorrow morning will be between 76 and 80 with partly cloudy skies \\
    \midrule
    \textit{User query} & - & - & - & - & - & On the 12th of this month would be great. \\
    \textit{Template} & - & - & - & - & - & Would you like to fly with American Airlines? The onward flight takes off at 4 am. It has a layover. The returning flight takes off at 12:45 pm. The ticket costs \$552 \\
    \textit{Reference} & - & - & - & - & - & How about a connecting American Airlines flight taking off at 4 am and costing \$552? The return time is at 12:45 pm. \\
    \textbf{Predictions} \\
    \textit{Baseline} & -0.004 & 2.50 & 2.17 & 0.83 & 0.0 & Okay. I've found an American Airlines flight departing at 4 am and returning at 12:45 pm. I inform you that the {\color{red}flight has a return flight at 12:45 pm}. The ticket is \${\color{red}1052}. \\
    \textit{Self-training} & 0.996 & 3.00 & 2.83 & 0.67 & 1.0 & I've found an American Airlines flight departing at 4 am and returning at 12:45 pm. This will cost you \$552. \\
    \textit{Full} & 0.998 & 2.00 & 2.00 & 0.50 & 1.0 & There is an American Airlines flight that leaves at 4 am and has a layover and a return flight at 12:45 pm for \$552. \\
    \bottomrule
    \end{tabular}
    \caption{Sample responses from different models on unseen test split for FewShotWeather (top row) and FewShotSGD (bottom row) datasets. We use 1shot-250 (FewShotWeather)/ 5-shot (FewShotSGD) train splits to fine-tune baseline and BLEURT based self-training. Grammaticality (Gram), naturalness (Nat) are on the scale of $1$ to $3$, informativeness (Info) is on the scale of $0$ to $1$ and accuracy (Acc) is binary. In general, we see that the baseline model generate responses that are repetitive in nature, contain {\color{red}novel content} and/or are missing relevant details.}
    \label{tab:qualitative}
\end{table*}

%% file: sections/05_relatedwork.tex
\section{Related Work}
\paragraph{Data-to-Text Generation}
While early research focused on rule-based  methods \cite{reiter2000building}, more recent work has relied heavily on neural methods  \citep{wen2015semantically,marcheggiani2018deep}.
Some recent works (\citet{kale2020text}, \citet{peng2020few}, \citet{kale2020machine}) showed that transfer learning from pre-trained language models can improve generalization capabilities and sample efficiency. In other lines of work, \citet{ferreira2019neural, moryossef2019step} find that pipelined neural approaches with explicit planning steps can outperform their end-to-end counterparts, while \citet{kale2020template} and \citet{du2020schema} showed the benefits of schema and template guided input representations. Inspired by \citet{kale2020template} we propose a simple and generic way to produce text-to-text representation, and study how it impacts compositional generalization.

\paragraph{Self-training for NLG} \citet{he2019revisiting} revisits the problem of self-training for NLG. They found that noise (from perturbing the input space) helps in self-training and propose a ``noisy'' version of self-training by augmenting vanilla training with the inputs from a reconstruction model. Building on this idea, the contemporary work \citep{heidari2021getting} on few-shot data-to-text generation proposes to self-train the model and shows efficacy on the Weather dataset. Another contemporary work \citep{li2021self} proposes to use constrained decoding to generate valid pseudo-responses for self-training and show convincing benefits. However, our work focuses on compositional generalization, rather than the pure few-shot learning setup. We propose a BLEURT-based self-training method, which is more generic than pseudo-response selection methods that rely on output structures.

%% file: sections/06_conclusion.tex
\section{Conclusion and Future Work}
We systematically study the problem of compositional generalization for data-to-text generation and show that existing state-of-the-art generation models do not generalize to unseen structures. We propose a simple and generic way to produce template guided text representation for response generation, and demonstrate its effectiveness on both seen and unseen structures. Further, we introduce a generic self-training approach that leverages fine-tuned BLEURT for pseudo response selection and show significant improvements over vanilla self-training on existing few-shot data-to-text generation benchmarks.

While our method requires only a small number of templates to start with, we still need to manually generate them for every unseen MR. Automatically generating templates by priming GPT-style models is an interesting line of future work. Furthermore, the effectiveness of our self-training method is highly dependent on the quality of the underlying BLEURT model (see Table \ref{tab:multiselftrain}). Given BLEURT based quality estimator is a learned model, it may be susceptible to data distribution shifts. We leave such analysis to future work. Another interesting future direction is to investigate the effectiveness of our approach to languages other than English.

%% file: sections/07_ethics_statement.tex
\section*{Ethics Statement}

To study compositional generalization for data-to-text tasks, we introduce data splits based on the already existing, publicly available, and widely used compositional weather dataset \citep{balakrishnan2019constrained}. We release our data splits to facilitate the development of new methods and consistent evaluation of them in comparison with existing works. In terms of use-case scenarios, we focus on task-oriented dialogue generation by using large pre-trained language models. These models are known to exhibit and potentially amplify social biases found in the training data, such as gender biases \citep{dinan2020queens}, and are capable of generating toxic or otherwise unsafe content \citep{weidinger2021ethical}. Our method helps these models generate higher quality responses than considered baselines when evaluated in terms of grammaticality, naturalness, informativeness, and accuracy. However, our work does not explicitly focus on mitigating social biases, unsafe content, or other potential ethical or social harms that might result from dialogue generation. Therefore, we caution against the deployment of our system in environments where any such biases can negatively impact the individuals interacting with our system without further assessment of the safety of this system in that environment.

%% file: sections/08_appendix.tex
\onecolumn
\section{Appendix}
\label{sec:appendix}

\begin{table*}[h!]
    \centering
    \small
    \begin{tabular}{p{1.4cm}rrrrrp{8.1cm}}
    \toprule 
    Fields & BLEURT & Gram & Nat & Info & Acc & Input or output response \\
    \midrule
    \textit{User query} & - & - & - & - & - & Will it be raining at 2 pm today \\
    \textit{Template} & - & - & - & - & - & No, and there won't be rain from 03:00 PM to 07:00 PM there will be partly cloudy with a temperature of 80 from 03:00 PM to 07:00 PM \\
    \textit{Reference} & - & - & - & - & - & No, rain isn't expected at 03:00 PM. Instead, expect partly cloudy skies and 80 fahrenheit continuing through 07:00 PM. \\
    \textbf{Predictions} \\
    \textit{Baseline} & 0.001 & 2.67 & 2.67 & 0.67 & 1.0 & No, it won't rain {\color{red} this} today from 03:00 PM to 07:00 PM and it will be 80 fahrenheit. Between 03:00 PM and 07:00 PM, you can expect partly cloudy skies{\color{red}, and 80 fahrenheit} \\
    \textit{Self-training} & 0.997 & 2.83 & 2.83 & 0.83 & 1.0 & No, it won't rain today from 03:00 PM to 07:00 PM. Between 03:00 PM and 07:00 PM, you can expect partly cloudy skies, and 80 fahrenheit \\
    \textit{Full} & 0.999 & 2.33 & 2.17 & 1.00 & 0.0 & No, it won't rain from 03:00 PM until 07:00 PM. It'll be 80 and partly cloudy \\
    \midrule
    \textit{User query} & - & - & - & - & - & What will the temperature be in Birmingham today? \\
    \textit{Template} & - & - & - & - & - & There will be with a temperature high of 69 low of 40 at today in Birmingham England United Kingdom \\
    \textit{Reference} & - & - & - & - & - & The high for today will be 69 and the low will be 40 fahrenheit in Birmingham, England, United Kingdom \\
    \textbf{Predictions} \\
    \textit{Baseline} & 0.000 & 2.17 & 2.33 & 0.83 & 1.0 & The temperatures in Birmingham, England, United Kingdom {\color{red} in Birmingham, England, United Kingdom} range between 40 and 69 \\
    \textit{Self-training} & 0.990 & 2.33 & 2.17 & 0.50 & 1.0 & The temperatures in Birmingham, England, United Kingdom today range between 40 and 69 \\
    \textit{Full} & 0.999 & 2.67 & 2.50 & 1.00 & 1.0 & There will be a high of 69 and low of 40 today in Birmingham, England, United Kingdom \\
    \midrule
    \textit{User query} & - & - & - & - & - & is there hail today \\
    \textit{Template} & - & - & - & - & - & No, but there will be heavy freezing rain and cloudy at today in Oxford \\
    \textit{Reference} & - & - & - & - & - & No, today in Oxford is cloudy with heavy freezing rain \\
    \textbf{Predictions} \\
    \textit{Baseline} & -0.001 & 2.33 & 2.67 & 0.83 & 0.0 & No, but there is {\color{red} cloudy hail} today in Oxford \\
    \textit{Self-training} & 0.999 & 3.00 & 3.00 & 0.50 & 1.0 & No, but there is cloudy weather today with heavy freezing rain in Oxford \\
    \textit{Full} & 0.998 & 2.33 & 2.50 & 0.83 & 1.0 & No, but it'll be cloudy with heavy freezing rain today in Oxford \\
    \midrule
    \textit{User query} & - & - & - & - & - & Should I wear a heavy jacket wednesday \\
    \textit{Template} & - & - & - & - & - & No there will be mostly sunny with a temperature high of 66 low of 47 at next Wednesday jacket is not recommended at next Wednesday, because there won't be cold at next Wednesday \\
    \textit{Reference} & - & - & - & - & - & No, you don't need a jacket, because it won't be cold next Wednesday. Expect a high of 66 and a low of 47, with mostly sunny skies \\
    \textbf{Predictions} \\
    \textit{Baseline} & -0.007 & 2.67 & 2.67 & 0.50 & 1.0 & No, it will not be cold, so you won't need a jacket. It will be mostly sunny with a high of 66 and a low of 47 next {\color{red}\st{Wednesday}} \\
    \textit{Self-training} & -0.001 & 1.83 & 1.83 & 0.83 & 1.0 & No, you won't need a jacket, it's not supposed to be cold next {\color{red}\st{Wednesday}}. It'll be mostly sunny with a high of 66 and a low of 47 \\
    \textit{Full} & 1.000 & 2.83 & 2.67 & 0.83 & 1.0 & No, you don't need a jacket, because it won't be cold next Wednesday. Expect a high of 66 and a low of 47, with mostly sunny skies \\
    \bottomrule
    \end{tabular}
    \caption{A few examples of sample responses from different models on unseen test split for FewShotWeather dataset. We use 1shot-250 train split to fine-tune baseline and BLEURT based self-training. Grammaticality (Gram), naturalness (Nat) are on the scale of $1$ to $3$, informativeness (Info) is on the scale of $0$ to $1$ and accuracy (Acc) is binary. In general, we see that the baseline model generate responses that are repetitive in nature and either contain {\color{red}novel content} or are {\color{red}\st{missing relevant details}}.}
    \label{tab:qualitative_weather}
\end{table*}

\begin{table*}
    \centering
    \small
    \begin{tabular}{p{1.5cm}rrrrrp{8cm}}
    \toprule 
    Fields & BLEURT & Gram & Nat & Info & Acc & Input or output response \\
    \midrule
    \textit{User query} & - & - & - & - & - & Find me the contact number and the cost per night. \\
    \textit{Template} & - & - & - & - & - & It costs \$93 per night. Their phone number is +44 20 7402 0281. \\
    \textit{Reference} & - & - & - & - & - & You can contact them on +44 20 7402 0281 and the cost per night is \$93. \\
    \textbf{Predictions} \\
    \textit{Baseline} & -0.003 & 2.67 & 3.00 & 0.17 & 0.0 & The phone number {\color{red} for Aslam's Rasoi in San Francisco} is +44 20 7402 {\color{red} 0181}. \\
    \textit{Self-training} & 0.998 & 2.83 & 2.83 & 0.67 & 1.0 & The cost is \$93 per day. The number is +44 20 7402 0281. \\
    \textit{Full} & 0.998 & 3.00 & 2.83 & 0.67 & 1.0 & It's \$93 per night and you can call them at +44 20 7402 0281. \\
    \midrule
    \textit{User query} & - & - & - & - & - & That sounds good. How much is the rental per day? \\
    \textit{Template} & - & - & - & - & - & Your car has been reserved. It costs \$32.00 per day. \\
    \textit{Reference} & - & - & - & - & - & It will cost you \$32.00 per day. Your car has been reserved. \\
    \textbf{Predictions} \\
    \textit{Baseline} & -0.004 & 2.33 & 2.00 & 0.67 & 0.0 & Your car has been reserved. It costs \${\color{red}32,32.00} per day. \\
    \textit{Self-training} & -0.002 & 3.00 & 2.83 & 0.67 & 0.0 & Your car has been reserved. It costs \${\color{red}32.50} per day. \\
    \textit{Full} & 0.998 & 3.00 & 3.00 & 0.67 & 1.0 & Your car has been reserved. The cost is \$32.00 per day. \\
    \midrule
    \textit{User query} & - & - & - & - & - & Yes, set up an appointment for me. \\
    \textit{Template} & - & - & - & - & - & What date would you like the appointment for? What is your preferred time for the appointment? \\
    \textit{Reference} & - & - & - & - & - & What date and time do you prefer? \\
    \textbf{Predictions} \\
    \textit{Baseline} & -0.003 & 3.00 & 3.00 & 0.83 & 1.0 & What date {\color{red}\st{and time}} would you like the appointment? \\
    \textit{Self-training} & 0.804 & 2.83 & 2.67 & 1.00 & 1.0 & What date would you like the appointment and what time do you want the appointment? \\
    \textit{Full} & 0.700 & 2.50 & 2.50 & 0.67 & 1.0 & What date and time would you like the appointment for? \\
    \midrule
    \textit{User query} & - & - & - & - & - & I want to find a property in Palo Alto. \\
    \textit{Template} & - & - & - & - & - & Do you want to find a place to rent or buy? \\
    \textit{Reference} & - & - & - & - & - & Are you looking to rent or buy? \\
    \textbf{Predictions} \\
    \textit{Baseline} & -0.004 & 2.67 & 2.67 & 0.67 & 1.0 & Do you want a house {\color{red}\st{for rent or purchase}}? \\
    \textit{Self-training} & 0.999 & 2.50 & 2.50 & 0.83 & 1.0 & Do you want a place for rent or purchase? \\
    \textit{Full} & 0.815 & 2.67 & 2.50 & 0.83 & 1.0 & Do you want to rent {\color{red}\st{or purchase}} a place? \\
    \midrule
    \textit{User query} & - & - & - & - & - & I also need to find a restaurant there so that I can eat something on that day. The restaurant I am interested in should be considered as reasonably priced, and that is serves Dumplings. \\
    \textit{Template} & - & - & - & - & - & I found 1 restaurants you may like. Hong Kong East Ocean Seafood Restaurant is a nice restaurant. It's located in Emeryville. \\
    \textit{Reference} & - & - & - & - & - & There is 1 restaurant which matches what you want. A great restaurant found in Emeryville is the Hong Kong East Ocean Seafood Restaurant. \\
    \textbf{Predictions} \\
    \textit{Baseline} & -0.003 & 2.17 & 2.33 & 0.83 & 0.0 & I have 1 option {\color{red}an} it is Hong Kong East Ocean Seafood {\color{red}\st{Cafe located in Emeryville}}. \\
    \textit{Self-training} & 0.857 & 2.17 & 2.50 & 0.83 & 1.0 & I have 1 option {\color{red} an} it is Hong Kong East Ocean Seafood Cafe located in Emeryville. \\
    \textit{Full} & 0.998 & 2.67 & 2.67 & 0.83 & 1.0 & I found 1 restaurant you might like. How about the Hong Kong East Ocean Seafood Restaurant in Emeryville? \\
    \bottomrule
    \end{tabular}
    \caption{A few examples of sample responses from different models on unseen test split for FewShotSGD dataset. We use 5-shot train split to fine-tune baseline and BLEURT based self-training. Grammaticality (Gram), naturalness (Nat) are on the scale of $1$ to $3$, informativeness (Info) is on the scale of $0$ to $1$ and accuracy (Acc) is binary. In general, we see that the baseline model generate responses that are incoherent, not factually grounded in the input, contain {\color{red}novel content} and/or are {\color{red}\st{missing relevant details}}.}
    \label{tab:qualitative_sgd}
\end{table*}

\begin{table*}[ht!]
    \centering
    \small
    \begin{tabular}{l|c|r|cc|cc}
    \toprule 
    Model & Self- & No. of & \multicolumn{4}{c}{FewShotSGD} \\
     & training & training & \multicolumn{2}{c}{Seen schemata} & \multicolumn{2}{c}{Unseen schemata} \\
     & iteration & examples & BLEU $\uparrow$ & SER $\downarrow$ & BLEU $\uparrow$ & SER $\downarrow$ \\
    \midrule
    Baseline & - & 558 & $20.66$ & $22.84$ & $20.52$ & $19.93$ \\
    \midrule
      \multirow{2}{*}{Vanilla} & 1 & + $101,577$ & $22.96$ & $16.26$ & $21.69$ & $15.19$ \\
       & 2 & + $30,867$ & $22.94$ & $15.43$ & $21.94$ & $16.04$ \\
       & 3 & + $5,998$ & $23.03$ & $15.15$ & $21.97$ & $15.96$ \\
      \midrule
      \multirow{2}{*}{BLEURT} & 1 & + $101,577$ & $24.34$ & $9.85$ & $23.29$ & $8.43$ \\
       & 2 & + $30,867$ & $24.84$ & $6.96$ & $23.64$ & $6.58$ \\
       & 3 & + $5,998$ & $25.22$ & $4.78$ & $24.13$ & $5.39$ \\
    \bottomrule
    \end{tabular}
    \caption{Model performance over multiple self-training iterations with 5-shot train split (FewShotSGD). $\uparrow$ indicates higher is better, $\downarrow$ indicates lower is better. We observe that model performance increases with the self-training iteration. However, the number of additional examples added decreases over iteration, suggesting that 2-3 iterations are sufficient for self-training.}
    \label{tab:multiselftrainsgd}
\end{table*}